\documentclass{article}
\usepackage{ijcai21}

\usepackage{times}
\usepackage{soul}
\usepackage{url}
\usepackage[hidelinks]{hyperref}
\usepackage[utf8]{inputenc}
\usepackage[small]{caption}
\usepackage{graphicx}
\usepackage{amsmath}
\usepackage{amsthm}
\usepackage{booktabs}
\usepackage{algorithm}
\usepackage{algorithmic}
\usepackage{dsfont}% \mathds{1}
\usepackage{amsfonts} % \mathbb
\usepackage{color}

\title{Interpretable Compositional Convolutional Neural Networks}

\author{
Wen Shen$^{1,*}$\and
Zhihua Wei$^{1,*}$\and
Shikun Huang$^1$\and
Binbin Zhang$^1$\and
Jiaqi Fan$^1$\and
Ping Zhao$^1$\and \\
Quanshi Zhang$^{2,\dag}$
\affiliations
$^1$Tongji University, Shanghai, China  \\
$^2$Shanghai Jiao Tong University, Shanghai, China \\
\emails
\{wen\_shen,zhihua\_wei,hsk,0206zbb,1930795,zhaoping\}@tongji.edu.cn,zqs1022@sjtu.edu.cn
}

\begin{document}

\maketitle

{
	\renewcommand{\thefootnote}{\fnsymbol{footnote}}
	\footnotetext[1]{Wen Shen and Zhihua Wei have equal contributions.}
	\footnotetext[2]{Quanshi Zhang is the corresponding author. He is with the John Hopcroft Center and the MoE Key Lab of Artificial Intelligence, AI Institute, at the Shanghai Jiao Tong University, China.}
}

\begin{abstract}
The reasonable definition of semantic interpretability presents the core challenge in explainable AI. This paper proposes a method to modify a traditional convolutional neural network (CNN) into an interpretable compositional CNN, in order to learn filters that encode meaningful visual patterns in intermediate convolutional layers. In a compositional CNN, each filter is supposed to consistently represent a specific compositional object part or image region with a clear meaning. The compositional CNN learns from image labels for classification without any annotations of parts or regions for supervision. Our method can be broadly applied to different types of CNNs. Experiments have demonstrated the effectiveness of our method. \emph{The code will be released when the paper is accepted.}
\end{abstract}

\section{Introduction}
Convolutional neural networks (CNNs) have exhibited superior performance in many visual tasks. Besides, the interpretability of CNNs has received increasing attention in recent years. Studies of network interpretability usually focus on the visualization of network features or the extraction of pixel-level correlations between network inputs and outputs. Training a CNN with interpretable features in intermediate layers is still a challenge to state-of-the-art algorithms, which helps people obtain more trustworthy and verifiable features.

\begin{figure}
	\centering
	\includegraphics[width=\linewidth]{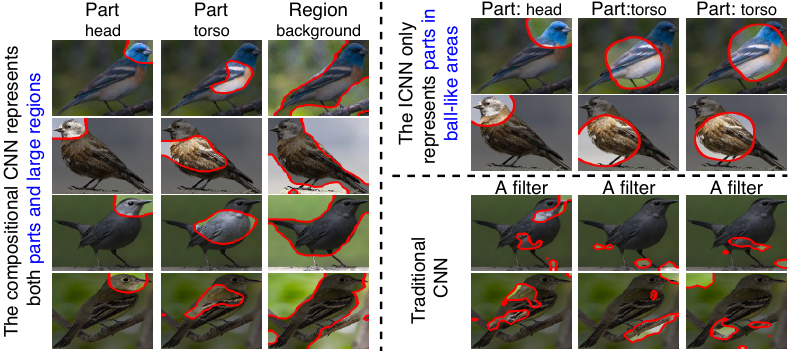}
	\caption{Compared with ICNN \protect\cite{zhang2018interpretable}, the interpretable compositional CNN defines the filter interpretability in a more generic manner, thereby modeling more diverse visual patterns. In a compositional CNN, each filter consistently represents a specific object part or image region through different images. Different filters represent different object parts and image regions. In comparison, the ICNN can only represent object parts in ball-like areas.}
	\label{fig:fig1-fjq}
\end{figure}

In this paper, we aim to propose a method to modify a CNN, which makes filters in an intermediate layer encode interpretable and compositional features. Specifically, as Fig.~\ref{fig:fig1-fjq} shows, each filter in the intermediate layer is supposed to be consistently activated by the same object part with specific shapes (\emph{e.g.} the head part of a bird) or the same image region without specific structures (\emph{e.g.} the sky in the background) through different images. Besides, different filters in the layer are supposed to be activated by different parts or regions, which ensures the diversity of visual patterns.

Given different images, we learn the interpretable compositional CNN in an end-to-end manner \textbf{without} any annotations of object parts or image regions for supervision. To this end, we add a specific loss to the intermediate layer in a CNN. This loss encourages each filter to be consistently activated by the same object part or the same image region. The loss also pushes different filters to be activated by different object parts or image regions.

We notice that a CNN usually uses a set of filters to jointly represent a specific object part or image region, instead of using a single filter, which has been discussed in \cite{fong2018net2vec}. Therefore, we divide filters in a convolutional layer into different groups. The loss is designed to force filters in the same group to be activated by the same object part or the same image region, and force filters in different groups to be activated by different parts or regions. Note that each filter in the group is required to represent almost the entire part/region, instead of a random sub-part/sub-region fragment inside, which ensures the clarity of the meaning of each filter. The mutual verification of visual patterns between filters in the same group ensures the stability of the visual patterns represented by each filter in this group. The slight difference of feature maps between filters in the same group encodes the fine-grained variety of the same type of parts/regions. To this end, we design a metric to measure the similarity between filters, which enables the loss to group filters. Besides, for multi-category classification, we design a loss to force different groups of filters to be activated by object parts or image regions of different categories.

In this study, we evaluate the interpretability of filters in the convolutional layer qualitatively and quantitatively. We visualize the feature map of a filter to qualitatively show the consistency of a filter's visual patterns through different images, in order to examine the fitness between the visual patterns automatically learned by a compositional CNN and the visual concepts in human's cognition. For the quantitative evaluation, previous metrics in \cite{zhang2018interpretable} can only evaluate semantic consistency of object parts in ball-like areas and strong priors of object structures. Therefore, we design two metrics to evaluate both the consistency of a filter's visual patterns and the diversity of visual patterns represented by different filters, respectively.

Previous studies also developed CNNs, where filters in an intermediate layer represented meaningful features. Capsule nets \cite{sabour2017dynamic} encoded different meaningful features, but these features usually did not represent parts or regions. \citeauthor{zhang2018interpretable} \shortcite{zhang2018interpretable} proposed interpretable CNNs (ICNNs), which learned filters in intermediate layers to represent object parts. They designed the information-theoretic interpretability loss to force filters to represent specific object parts. Filters in the ICNN could only represent object parts in ball-like areas. In comparison, we extend the filter interpretability to both object parts with specific shapes and image regions without clear structures, which proposes significant challenges to state-of-the-art algorithms. Thus, the compositional CNN can encode more types of features than the ICNN. Please see Fig.~\ref{fig:fig1-fjq} for details.

Contributions of this study can be summarized as follows. We propose a method to modify traditional CNNs into compositional CNNs without any annotations of object parts or image regions for supervision. Each filter in a compositional CNN consistently represents the same object part or image region with a clear meaning. Experiments show that our method can be broadly applied to CNNs with different architectures.

\section{Related Work}

\paragraph{Learning interpretable features.} Some studies directly trained networks to increase the interpretability of intermediate-layer features. Capsule nets \cite{sabour2017dynamic} learned capsules to encode meaningful features via a dynamic routing mechanism. InfoGAN \cite{chen2016infogan} and {\small$\beta$}-VAE \cite{higgins2017beta} learned interpretable representations for generative networks. These studies did not make each filter in the CNN encode a specific visual pattern. To this end, some studies \cite{liinterpretable,liang2020training} learned class-specific filters, \emph{i.e.} each filter only represented a specific category. However, such class-specific filters could not represent fine-grained meaningful visual patterns, such as object parts and image regions. \citeauthor{chen2019looks} \shortcite{chen2019looks} proposed the ProtoPNet to extract similar object-part regions that were shared for fine-grained classification. The ProtoPNet did not ensure each filter to represent a clear meaning. \citeauthor{zhang2018interpretable} \shortcite{zhang2018interpretable} proposed interpretable CNNs to make each filter in a high convolutional layer represent a specific object part. In comparison, we extend the filter interpretability to both object parts and image regions, which presents significant challenges to state-of-the-art algorithms.

\paragraph{Compositional models.} Previous studies in compositional models focused on learning hierarchical feature representations \cite{fidler2007towards,zhu2010part,ommer2009learning}, such as graph-based models \cite{si2013learning,wang2015semantic} and part-based models \cite{ott2011shared,zhu2010latent}. These models did not use neural networks to learn features. Other studies learned discriminative compositional parts directly through network training. \citeauthor{stone2017teaching} \shortcite{stone2017teaching} manually designed a graphical model to organize CNN modules and represent object structures. \citeauthor{kortylewski2020compositional} \shortcite{kortylewski2020compositional} designed a specific compositional layer to enable the network to localize partial occlusion. \citeauthor{Huang_2020_CVPR} \shortcite{Huang_2020_CVPR} learned discriminative object parts for fine-grained recognition based on manually labeled part priors. However, in all above studies, the compositional information of features was not automatically learned from data. In comparison, the proposed compositional CNN automatically learns compositional features without any annotations of parts or regions. \emph{I.e.} the compositional CNN automatically regularizes its features into meaningful parts and regions without letting people supervise its semantic representations.

\section{Algorithm}

In this section, we aim to modify a convolutional layer of a CNN into an interpretable compositional layer. In this layer, each filter is supposed to consistently represent the same object part or the same image region through different images. To ensure the consistency of the visual patterns represented by each filter, we use a group of filters to represent the same object part or the same image region. The set of filters {\small$\Omega=\{1,2,\cdots,d\}$} in the target layer are divided into different groups {\small$A_1,A_2,\cdots,A_K$}, where {\small$A_1\cup A_2\cup \cdots\cup A_K=\Omega$}; {\small$A_i\cap A_j=\emptyset$}. {\small$\mathbf{A}=\{A_1,A_2,\cdots,A_K\}$} denotes the partition of filters. Let {\small$\theta$} denote parameters of the CNN. Given a set of training images, we aim to simultaneously optimize parameters {\small$\theta$} and the partition {\small$\mathbf{A}$} to ensure that filters in the same group consistently represent the same visual patterns through different images, and filters in different groups represent different visual patterns.

To measure the similarity of visual patterns represented by different filters, we propose a metric to measure the similarity between filters. Given an image {\small$I$}, let {\small$x_i^I\in\mathbb{R}^{m}$} denote the feature map of the {\small$i$}-th filter in the target convolutional layer after the ReLU operation. Given a set of {\small$n$} training images {\small$\mathbf{I}$}, let {\small$X_i=\{x_i^I\}_{I\in\mathbf{I}}$} denote the set of feature maps of the {\small$i$}-th filter. Then, we compute the similarity between the {\small$i$}-th and the {\small$j$}-th filters, which represents whether these two filters consistently represent the same visual patterns through different images. This similarity is formulated as {\small$s_{ij}=\mathcal{K}(X_i,X_j)\in\mathbb{R}$}, where {\small$\mathcal{K}$} is a kernel function. Based on the similarity metric, we design the following loss to learn filters.
\begin{small}
	\begin{equation}\label{l_s_1}
	{\rm Loss}(\theta,\mathbf{A})
	=-\sum_{k=1}^{K}\frac{S_k^{\rm within} }{S_k^{\rm all}} = -\sum_{k=1}^{K}\frac{\sum_{i,j\in A_k} s_{ij} }{\sum_{i\in A_k,j\in \Omega}s_{ij}},
	\end{equation}
\end{small}
where {\small$S_k^{\rm within}=\sum_{i,j\in A_k} s_{ij}=\sum_{i,j\in A_k}\mathcal{K}(X_i,X_j)$} measures the similarity between filters \emph{within} the same group {\small$A_k$}; {\small$S_k^{\rm all}=\sum_{i\in A_k,j\in \Omega}s_{ij}=\sum_{i\in A_k,j\in \Omega} \mathcal{K}(X_i,X_j)$} measures the similarity between filters in {\small$A_k$} and \emph{all} filters in {\small$\Omega$}. This loss increases {\small$S_k^{\rm within}$} to ensure that filters in the same group have high similartiy, and decreases {\small$S_k^{\rm all}$} to ensure that filters in different groups have low similartiy. Specifically, the similarity metric is implemented as a kernel function.
\begin{small}
	\begin{equation}\label{sij}
	s_{ij}=\mathcal{K}(X_i,X_j)
	=\rho_{ij}+ 1
	=\frac{{\rm cov}(X_i,X_j)}{\sigma_i \sigma_j} +1 \ge 0,
	\end{equation}
\end{small}
where {\small$\rho_{ij}\in [-1,1]$} denotes the Pearson's correlation coefficient between variables {\small$x_i^I$} and {\small$x_j^I$} through different images; Constant 1 is added to ensure the non-negativity of the similarity. {\small${\rm cov}(X_i,X_j)\in\mathbb{R}$} denotes the covariance between variables {\small$x_i^I$} and {\small$x_j^I$} through different images, {\small${\rm cov}(X_i,X_j)=\frac{1}{n-1}\sum_{I\in\mathbf{I}}(x_i^I-\mu_i)^{\top}(x_j^I-\mu_j)\in\mathbb{R}$}; {\small$\mu_i=\frac{1}{n}\sum_{I\in\mathbf{I}}x_i^I\in\mathbb{R}^{m}$}; {\small$\sigma_i^2 = \frac{1}{n-1}\sum_{I\in\mathbf{I}}(x_i^I-\mu_i)^{2}\in \mathbb{R}$}. The similarity metric can be understood as the sum of similarity between feature maps of the {\small$i$}-th and the {\small$j$}-th filters through all training images as follows, {\small$s_{ij}=\mathcal{K}(X_i,X_j)=\sum_{I\in\mathbf{I}} \phi(x_i^I)^{\top} \phi(x_j^I)$}, where {\small${\phi(x_i^I)}^{\top}=[(x_i^I-\mu_i)^{\top},\sqrt{1-1/n}\sigma_i^{\top}] /\sqrt{n-1}\sigma_i$}.

The proposed loss makes filters in the same group have similar feature maps, which ensures the clarity and the stability of the visual patterns represented by each filter in this group. Meanwhile, the slight difference of these feature maps encodes fine-grained variety of the same type of parts/regions. Besides, the loss also makes filters in different groups have different feature maps, which ensures the diversity of the visual patterns represented by different groups of filters.

\paragraph{Binary classification of a single category.} We train a compositional CNN in an end-to-end manner by minimizing the following objective function.
\begin{small}
	\begin{equation}\label{loss_1}
	{\mathbf{L}}(\theta,\mathbf{A})=  \lambda \textbf{\rm Loss}(\theta,\mathbf{A}) +\frac{1}{n}{\sum}_{I\in\mathbf{I}}\textbf{\rm L}^{\rm cls}(\hat{y}_I,y_I^*;\theta),
	\end{equation}
\end{small}
where {\small$\textbf{\rm L}^{\rm cls}(\hat{y}_I,y_I^*;\theta)$} denotes the classification loss on image {\small$I$}; {\small$\hat{y}_I,y_I^*\in\{-1,+1\}$} denote the output of the CNN and the ground-truth label, respectively; {\small$\lambda$} is a positive weight.

\paragraph{Multi-category classification.} For the multi-category classification, besides {\small$\textbf{\rm Loss}(\theta,\mathbf{A})$}, we design another loss to make different groups of filters to be activated by parts or regions of different categories. Given a set of {\small$n$} training images {\small$\mathbf{I}$}, let {\small$\mathbf{I}_c\subset \mathbf{I}$} represent the subset of images of the category {\small$c$}, ({\small$c = 1, 2, \cdots , C$}). Filters in a certain group are supposed to be mainly activated by a specific object part or image region of very few categories, and keep silent on other categories. To this end, for each filter, we quantify the distribution of its neural activations over different categories. We propose a metric to measure the similarity between such distributions of different filters. Given the {\small$p$}-th image {\small$I$}, let {\small$z_k^{(p)}\in\mathbb{R}$} denote the average activation score of filters in group {\small$A_k$}, {\small$z_k^{(p)}=\frac{1}{|A_k|\cdot m}\sum_{i\in A_k}\sum_{u=1}^{m} x_{i,u}^I$}, where {\small$|A_k|$} denotes the number of filters in group {\small$A_k$}; {\small$x_{i,u}^I$} denotes the {\small$u$}-th element in {\small$x_i^I\in\mathbb{R}^m$}. The similarity between distributions of neural activations of different groups on the {\small$p$}-th and the {\small$q$}-th images is computed using the following kernel function. {\small$s_{pq}=\mathcal{K}(\mathbf{z}^{(p)},\mathbf{z}^{(q)})={(\mathbf{z}^{(p)}})^{\top}(\mathbf{z}^{(q)})\in\mathbb{R}$}, where {\small$\mathbf{z}^{(p)}=[z_1^{(p)},\cdots,z_K^{(p)}]^{\top}\in\mathbb{R}^K$}; {\small$x_{i,u}^I\ge 0$}, thereby {\small$s_{pq}\ge0$}. We propose the following loss to learn filters.
\begin{small}
	\begin{equation}\label{l_s_2}
	{\rm L}^{\rm multi}(\theta)  
	=-\sum_{c=1}^{C}\!\!\frac{\sum\limits_{p,q\in \textbf{I}_c} \!\!\!s_{pq}}{\sum\limits_{p\in \textbf{I}_c,q\in\mathbf{I}}\!\!\!s_{pq}} 
	= -\sum_{c=1}^{C}\frac{\sum\limits_{p,q\in \textbf{I}_c} \!\!\!\mathcal{K}(\mathbf{z}^{(p)},\mathbf{z}^{(q)})}{\sum\limits_{p\in \textbf{I}_c,q\in\mathbf{I}}\!\!\!\mathcal{K}(\mathbf{z}^{(p)},\mathbf{z}^{(q)})}.
	\end{equation}
\end{small}
The final objective function for multi-category classification is given as follows.
\begin{small}
	\begin{equation}\label{loss_2}
	{\mathbf{L}}(\theta,\mathbf{A}) = \lambda \textbf{\rm Loss}(\theta,\mathbf{A}) + \beta \textbf{\rm L}^{\rm multi}(\theta)+  \frac{1}{n}\sum_{I\in\mathbf{I}}\textbf{\rm L}^{\rm cls}(\hat{y}_I,y_I^*;\theta),
	\end{equation}
\end{small}
where {\small$\lambda$} and {\small$\beta$} are positive weights.

\paragraph{Learning.} During the learning process, we need to simultaneously optimize network parameters {\small$\theta$} and the filter partition {\small$\mathbf{A}$}. Fortunately, we find that, when we fix {\small$\theta$}, the minimization of {\small${\rm Loss}(\theta,\mathbf{A})$} \emph{w.r.t.} {\small$\mathbf{A}$} is essentially equivalent to the problem of the spectral clustering in \cite{shi2000normalized}. \emph{I.e.} we can rewrite {\small${\rm Loss}(\theta,\mathbf{A})$} as the following equation, which is exactly the same objective function in \cite{shi2000normalized}.
\begin{small}
	\begin{equation}\label{sc}
	\frac{1}{2}(\textbf{\rm Loss}(\theta,\mathbf{A}) + K)
	=\frac{1}{2}\sum_{k=1}^{K} \frac{ \sum_{i\in A_k,j\not\in A_k} s_{ij} }{\sum_{i\in A_k,j\in \Omega} s_{ij} }.
	\end{equation}
\end{small}
Here, we regard the set of filters {\small$\Omega$} as data points in the spectral clustering that need to be clustered into different groups {\small$A_1,\cdots,A_K$}. {\small$s_{ij}$} corresponds to the similarity between two data points. In this way, {\small$\mathbf{A}$} can be optimized by applying the clustering technique in \cite{shi2000normalized}. Therefore, we alternately optimize {\small$\theta$} and {\small$\mathbf{A}$} to minimize {\small${\rm Loss}(\theta,\mathbf{A})$}.

\section{Experiments}

We applied our method to CNNs with six types of architectures to demonstrate the broad applicability of our method. We used object images in four different benchmark datasets to learn compositional CNNs for both the binary classification of a single category and the multi-category classification. We designed two metrics to measure the inconsistency of a filter's visual patterns and the diversity of visual patterns represented by different filters. We also visualized feature maps of a filter to qualitatively show the consistency of a filter's visual patterns. We compared the performance of learning interpretable filters in different convolutional layers of a compositional CNN. We also discussed the effects of the group number on the performance of learning interpretable filters. For binary classification of a single category, we set $\lambda=1.0$ for most DNNs except for VGG-16 with $\lambda=0.1$. It was because the learning of a residual network could be considered as the optimization of massive parallel shallow networks. From this perspective, the VGG-16 was the most difficult to optimize. For multi-category classification, we set {\small$\lambda=0.1$} and {\small$\beta=0.1$}, because {\small$\textbf{\rm L}^{\rm multi}$} has partially taken the work of {\small$\textbf{\rm Loss}(\theta,\mathbf{A})$}. During the training procedure, for each time we optimized {\small$\theta$} through all training samples, we optimized {\small$\mathbf{A}$} once.

\begin{figure*}[tbp]
	\centering
	\includegraphics[width=\linewidth]{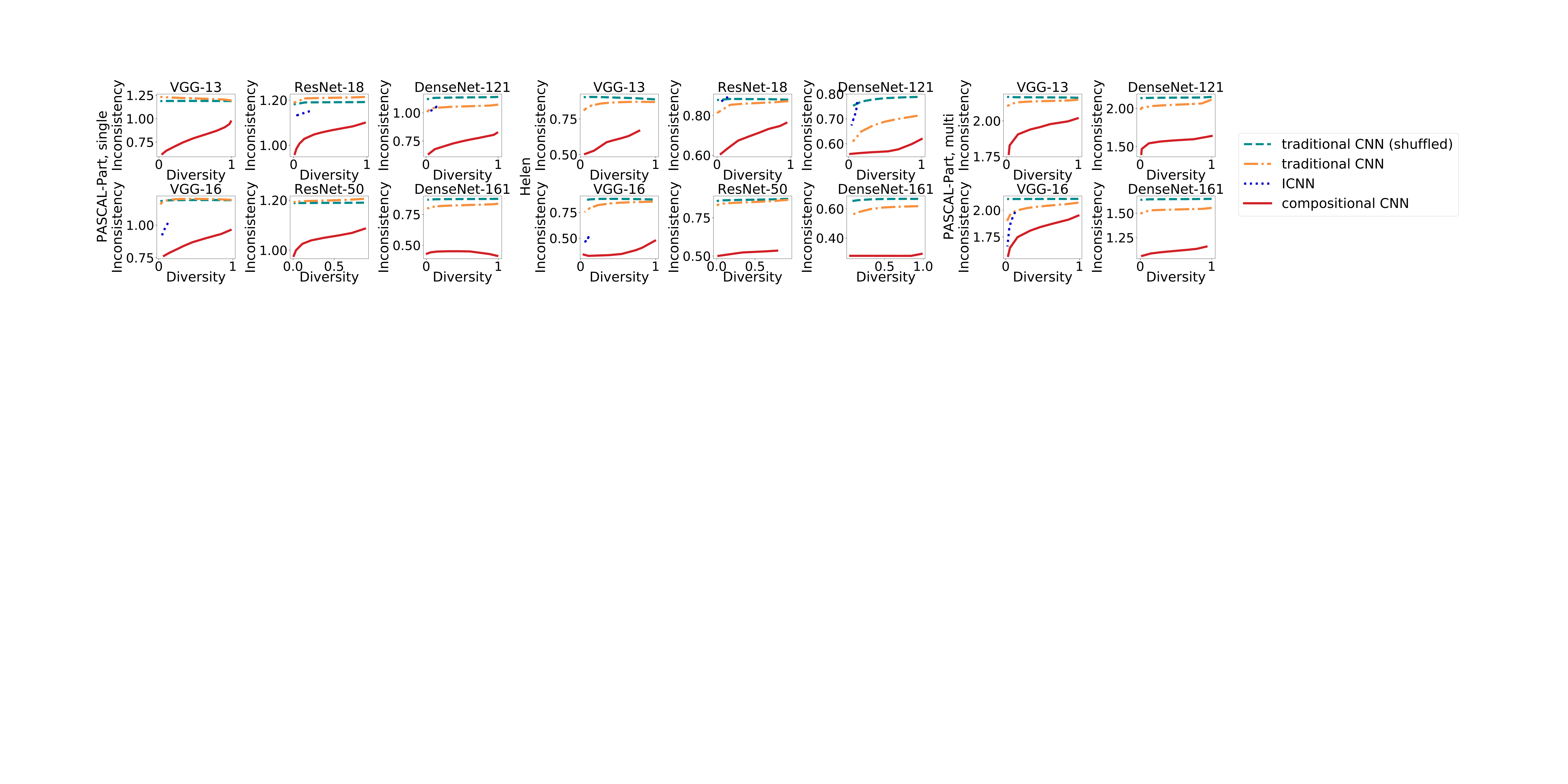}
	\includegraphics[width=0.6\linewidth]{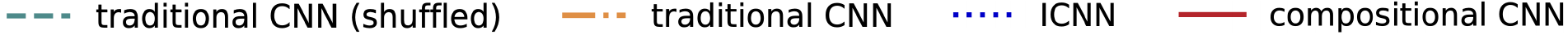}
	\caption{Comparisons of inconsistency of visual patterns and diversity of visual patterns between CNNs. For the binary classification of a single category, we showed curves of the average inconsistency of visual patterns and the average diversity of visual patterns over each CNN learned for each individual category. Results for each single category are shown in Fig.~\ref{fig:fig7}. Note that each value of inconsistency in this figure indicates the average inconsistency of all filters of a DNN.}
	\label{fig:metrics}
\end{figure*}

\subsection{Learning Compositional CNNs}
\subsubsection{Binary Classification of A Single Category} We learned six types of compositional CNNs based on the VGG-13\footnote{The VGG-13 and VGG-16 used in this paper were slightly revised by adding the batch-normalization \cite{BN2015} operation after each convolution layer.}, VGG-16\footnotemark[1] \cite{vgg}, ResNet-18, ResNet-50 \cite{he2016deep}, DenseNet-121, and DenseNet-161 \cite{huang2017densely} architectures. Just like in \cite{zhang2018interpretable}, we added the loss {\small$\textbf{\rm Loss}(\theta,\mathbf{A})$} to the high convolutional layer in a CNN. It was because the previous study \cite{bau2017network} had revealed that filters in high convolutional layers were more likely to represent object parts or image regions, instead of detailed patterns (\emph{e.g.} colors or textures). For the VGG-13, VGG-16, DenseNet-121, and DenseNet-161, we added the proposed loss to the top convolutional layer. For the ResNet-18, we added the loss to layer \emph{conv4\_4}. For the ResNet-50, we added the loss to layer \emph{conv4\_18}. All these compositional CNNs were learned based on the CUB200-2011 dataset \cite{wah2011caltech}, the Large-scale CelebFaces Attributes (CelebA) dataset \cite{liu2015faceattributes}, the Helen Facial Feature dataset \cite{smith2013exemplar}, and animal categories in the PASCAL-Part dataset \cite{chen2014detect}. In the field of learning interpretable deep features, animal categories were widely used to evaluate the automatically learned interpretable features \cite{zhang2018interpretable}. It was because animals usually contained deformable parts, which presented great challenges for part or region localization. Note that, the Helen Facial Feature dataset was usually used for the facial landmark localization. However, in this study, we used this dataset for the classification of faces and non-faces. It was because this dataset provided segmentation masks for face parts to evaluate the inconsistency and the diversity of visual patterns. We randomly selected the same number of samples from the PASCAL-Part dataset as negative samples for training and testing.

We followed experimental settings in \cite{zhang2018interpretable} to learn compositional CNNs for binary classification of a single category on the CUB200-2011 dataset and the PASCAL-Part dataset. For compositional CNNs learned from the CUB200-2011 dataset, the PASCAL-Part dataset, and the Helen Facial Feature dataset, we set {\small$K=5$}. For the CelebA dataset, we set {\small$K=16$}, because these compositional CNNs usually learned detailed visual patterns from face images.

To compare the performance of learning interpretable filters in different convolutional layers, we learned two compositional CNNs based on the VGG-16 architecture by adding the proposed loss to layer \emph{conv4\_3} and layer \emph{conv5\_3}, respectively. These two compositional CNNs were learned on the PASCAL-Part dataset. To explore the effects of different values of {\small$K$}, we learned two compositional CNNs based on the ResNet-50 architecture using the CelebA dataset by setting {\small$K=8$} and {\small$K=16$}, respectively.

\subsubsection{Multi-Category Classification} We learned four types of compositional CNNs based on the VGG-13, VGG-16, DenseNet-121, and DenseNet-161 architectures for the classification on the PASCAL-Part dataset following experimental settings in \cite{zhang2018interpretable}. We set {\small$K=16$}.

For all compositional CNNs, we learned traditional CNNs based on the same architectures and datasets as baselines. We replaced the zero padding with the replication padding for all compositional CNNs. For traditional CNNs based on the DenseNet architectures, we  initialized parameters of the fully-connected layers, and loaded parameters of other layers from the same architectures that were pre-trained using the ImageNet dataset \cite{DengImageNet}. For traditional CNNs based on other architectures, we initialized parameters of the target layer (\emph{i.e.} the convolutional layer would be modified to an interpretable compositional layer) and its following layers, and loaded parameters of other layers from the same architectures that were pre-trained using the ImageNet dataset. For all compositional CNNs, we loaded parameters of all layers from the above well-trained traditional CNNs.

\begin{figure*}
	\centering
	\includegraphics[width=0.9\linewidth]{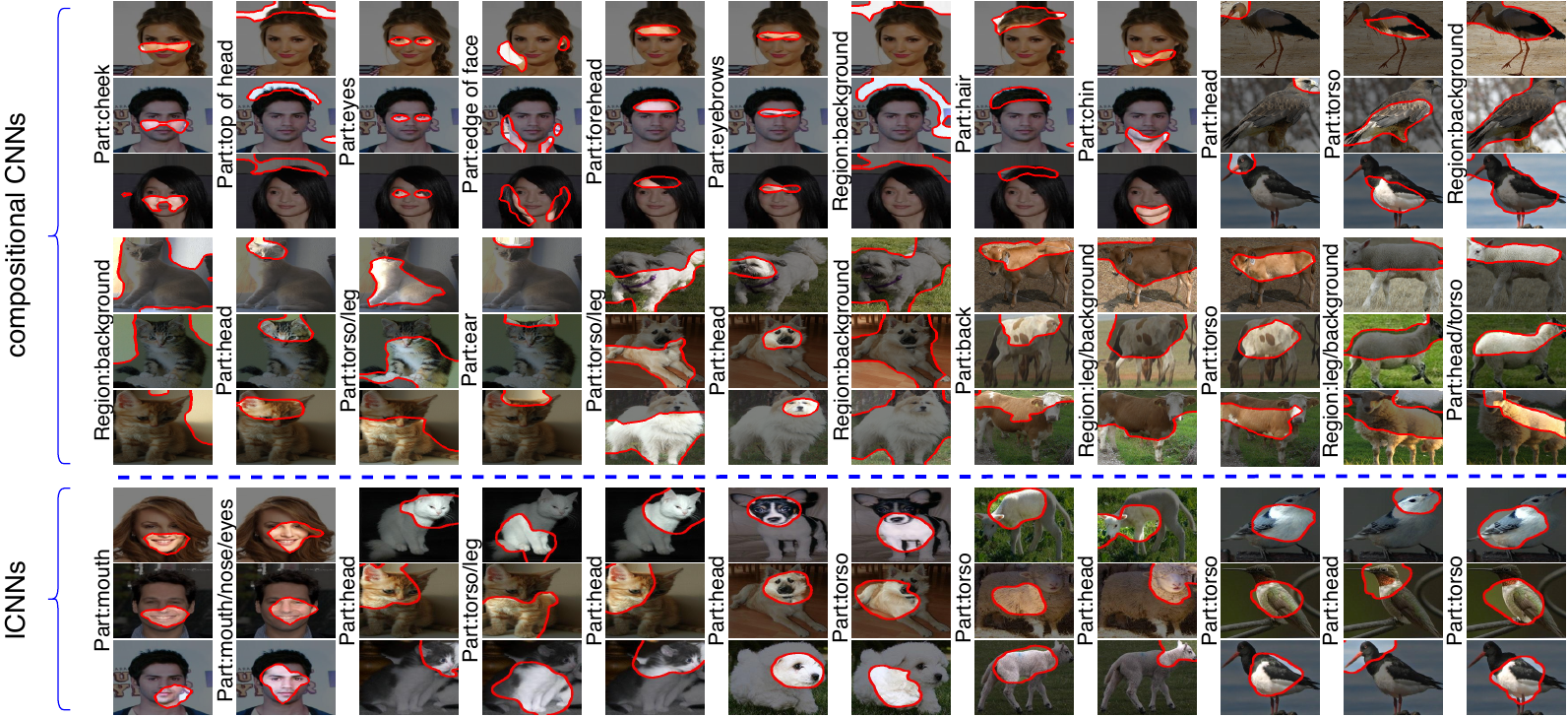}
	\caption{Visualization of feature maps of compositional CNNs and ICNNs \protect\cite{zhang2018interpretable}. Each column in the figure corresponds to a certain filter. Visualization results indicate that each filter in a compositional CNN consistently represented the same object part or the same image region, while different filters represented different parts and regions. In comparison, filters in an ICNN could only represent object parts. Note that we manually classified filters into part filters and region filters to help understand the visual patterns represented by the filter. In addition, part filters in the compositional CNN usually encoded more complex shapes than those in the ICNN.}
	\label{fig:fig2-fjq2}
\end{figure*}

\subsection{Quantitative Evaluation of Filter Interpretability}
Some previous studies also focused on learning interpretable filters, but their metrics usually have strong limitations and can not be used in our experiments. Metrics in \cite{zhang2018interpretable} can only be used to evaluate semantic consistency of object parts in ball-like areas and strong priors of object structures. \citeauthor{bau2017network} \shortcite{bau2017network} annotated six types of visual semantics for evaluation (including colors and materials), but filters in the compositional CNN were not designed towards such semantics. Therefore, we extended the metric in \cite{bau2017network} and proposed the inconsistency of visual patterns to evaluate the interpretability of filters. Besides, we evaluated the diversity of visual patterns represented by filters, which was an significant factor neglected in previous studies.

\subsubsection{Evaluation Metric 1: Inconsistency of Visual Patterns} This metric was proposed to measure the consistency of visual patterns represented by a filter through different images. Ideally, an interpretable filter was supposed to have high consistency. We computed the probability of a filter being associated with a ground-truth semantic concept in a specific image (\emph{e.g.} bird head, bird torso). Then, we defined the inconsistency of visual patterns as the entropy of such probabilities over different semantic concepts.

For simplicity, here, we only discussed the metric for a single filter below. We first computed the pixel-wise receptive field (RF) of neural activations of the filter on testing images \cite{zhang2018interpretable}. Let {\small$Q(I)\in\mathbb{R}^M$} denote activation scores of the target filter projected onto the test image {\small$I$}, where {\small$M$} denoted the number of pixels in the image {\small$I$}. We only considered activation scores in the feature map greater than a threshold {\small$\tau$} as valid ones to represent the filter (the setting of {\small$\tau$} would be explained later). Then, {\small$\tilde{Q}(I)\in\{0,1\}^M$} s.t. {\small$\tilde{Q}_u(I)=\mathds{1}(Q_u(I)\ge \tau)$} denoted the RF. Let {\small$G^j(I)\in\{0,1\}^M$} denote the ground-truth segmentation mask of the {\small$j$}-th concept ({\small$j=1,\cdots, T$}) on the test image {\small$I$}. The probability of the target filter being associated with the {\small$j$}-th concept was computed as {\small$P_j=\frac{\sum_{I\in\mathbf{I}^{\rm test}}\sum_{u=1}^M \min\{\tilde{Q}_u(I), G_u^j(I) \} }{\sum_{I\in\mathbf{I}^{\rm test}}\sum_{u=1}^M \tilde{Q}_u(I)}$}, where {\small$\mathbf{I}^{\rm test}$} denoted the set of testing images. Then, the inconsistency of the target filter's visual patterns was defined as the entropy {\small$H = -\sum_{j=1}^{T} P_j \log P_j$}.

\begin{figure*}
	\centering
	\includegraphics[width=\linewidth]{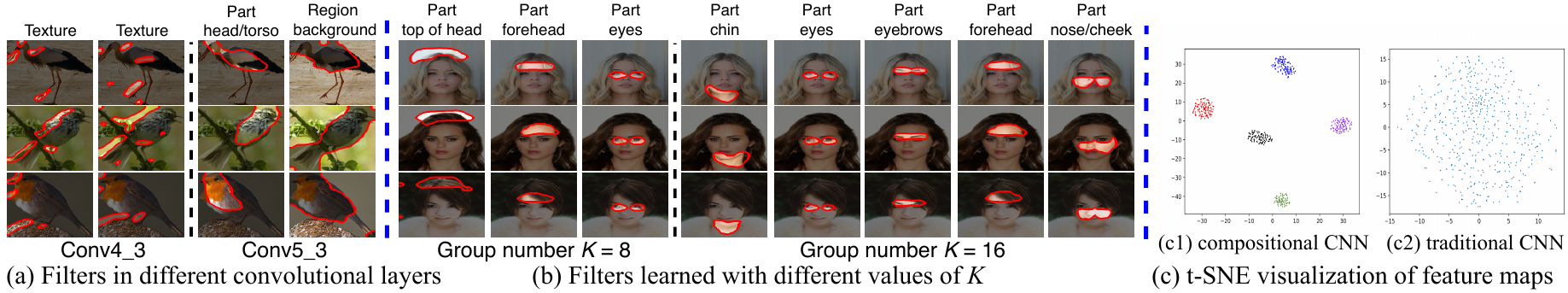}
	\caption{(a) Comparisons of interpretable filters in different convolutional layers. Results indicate that filters in a high convolutional layer tended to represent parts or regions, while filters in a middle convolutional layer tended to represent textures. (b) Filters learned with different values of {\small$K$}. Filters in the compositional CNN with {\small$K=16$} represented more detailed visual patterns than the CNN learned with {\small$K=8$}. (c) t-SNE visualization of feature maps of a compositional CNN (c1) and a traditional CNN (c2). Each point represents a feature map. Different colors of points represent feature maps of filters in different groups.}
	\label{fig:fig4}
\end{figure*}

\begin{figure}[tbp]
	\centering
	\includegraphics[width=\linewidth]{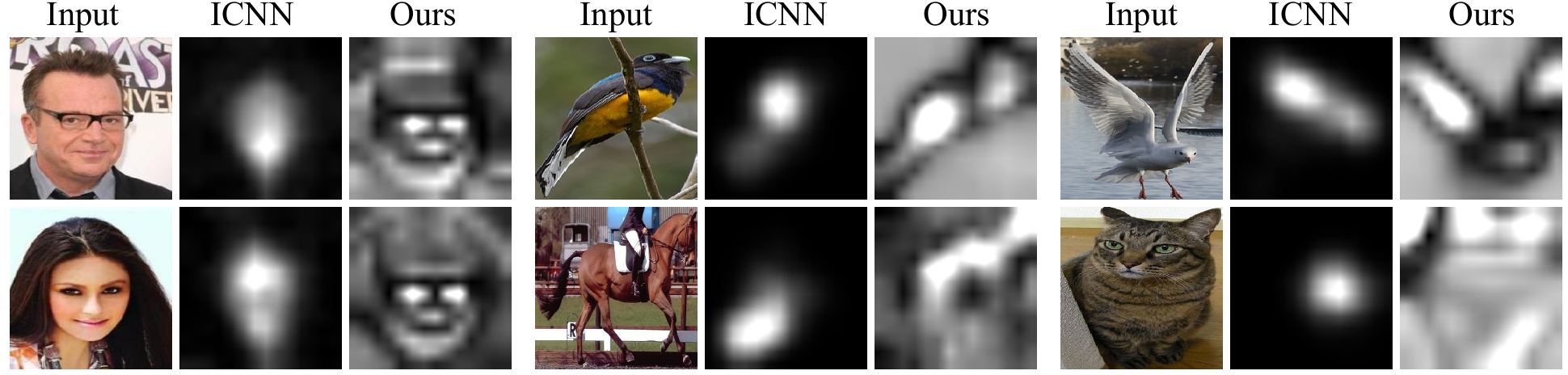}
	\caption{Visualizing distributions of visual patterns that are encoded in interpretable filters via the method in \protect\cite{zhang2018interpretable}. Results show that interpretable filters of a compositional CNN explained much more regions in an image than those of an ICNN.}
	\label{fig:mean_fmap}
\end{figure}

\begin{figure}[tbp]
	\centering
	\includegraphics[width=\linewidth]{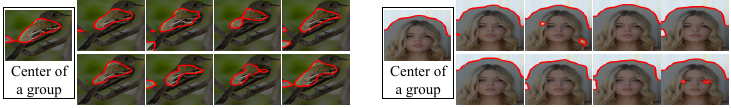}
	\caption{Comparisons of RFs between the center of the group and each filter in the group.}
	\label{fig:group}
\end{figure}

\paragraph{Binary classification of a single category.} We followed \cite{zhang2018interpretable} to merge certain areas of each animal category in the PASCAL-Part dataset to obtain stable landmark locations as stable concepts for evaluation. We used five concepts for the \emph{bird} category, including (head, l/r-eyes, beak, neck), (torso, l/r-wings), (l/r-legs/feet), (tail), and (background). Here, all parenthesized areas were merged as a new concept. We used four concepts for the \emph{cat} category, including (head, l/r-eyes, l/r-ears, nose, neck), (torso, tail), (lf/rf/lb/rb-legs, lf/rf/lb/rb-paws), and (background). We used four concepts for the \emph{dog} category, areas of which were merged in the same way as the cat category, except for merging the additional muzzle area to the head concept. We used four concepts for the \emph{cow} category, which were defined in a similar way as the dog category. We added l/r-horn to the head concept. We used four concepts for \emph{sheep} and \emph{horse} categories, which were defined in the same way as the cow category. Note that, in actual calculations, we only used images with relatively complete areas of each animal category\footnote{The dataset for the computation of metrics in this paper will be released in https://github.com/ada-shen/icCNN.}. In the Helen Facial Feature dataset\footnotemark[2], we used three concepts for the face category. We merged areas of face skin, l/r-eyebrow, l/r-eye, nose, u/l-lip, and inner mouth as the face concept. We used the areas of hair and background as the 2-nd and the 3-rd concepts, respectively.

\paragraph{Multi-category classification.} In the PASCAL-Part dataset, for each category, we used the foreground object as a single concept, and used the background as another concept. We considered visual concepts of all categories equally, \emph{i.e.} we would get {\small$T=2C$} concepts for the classification of {\small$C$} categories. Then, we used the aforementioned entropy {\small$H$} over the {\small$2C$} concepts for evaluation. Note that, for the classification of a large number of categories, theoretically, each category only obtained very few filters, which decreased the filter interpretability. Therefore, we only learned compositional CNNs for multi-category classification based on all animal categories in the PASCAL-Part dataset.

\paragraph{Randomly shuffled feature maps as baselines.} We constructed feature maps that totally had no consistency of visual patterns as a baseline. In implementation, we randomly shuffled different images' feature maps of a traditional CNN to approximately construct random feature maps.

\subsubsection{Evaluation Metric 2: Diversity of Visual Patterns} This metric was proposed to evaluate whether a CNN learned various visual patterns. In this study, the diversity of visual patterns was approximately quantified as the number of pixels which had been explained by a CNN. We determined that a pixel was explained by a CNN, if this pixel was explained by some filters. Recall that, we had computed the pixel-wise RF of neural activations of a filter on the test image {\small$I$} based on \cite{zhang2018interpretable}. Here, we used {\small$\tilde{Q}^i(I)\in\{0,1\}^M$} to denote the RF of neural activations of the {\small$i$}-th filter. Then, we determined that the {\small$u$}-th pixel was explained by a CNN, if {\small$(\frac{1}{d}\sum_{i=1}^d\tilde{Q}_u^i(I)) \ge\gamma$}. We set {\small$\gamma=0.2$}. The higher diversity meant that RFs of filters covered more pixels, \emph{i.e.} more diverse concepts were encoded by the CNN. Therefore, the diversity of visual patterns was computed as {\small$Diversity=\frac{1}{M}\mathbb{E}_{I}[\sum_{u=1}^{M}\mathds{1}((\frac{1}{d}\sum_{i=1}^d\tilde{Q}_u^i(I)) \ge\gamma)]$}.

\begin{table}[tbp]
	\centering
	\resizebox{\linewidth}{!}{
		\begin{tabular}{c|ccc|c}
			\toprule
			&\multicolumn{3}{c|}{ single-category} &	multi-category \!\!\!\! \\
			\cline{2-5}
			\!\!\!\!&\!\!\!\! PASCAL-Part     			\!\!\!\!&\!\!\!\! CUB200		\!\!\!\!&\!\!\!\! CelebA 		\!\!\!\!&\!\!\!\! PASCAL-Part \!\!\!\!\\
			\midrule
			%\hline
			VGG-13 &  {\bf97.07} &	{\bf99.76}	& --	 &   {\bf87.51} \\
			compositional CNN & 96.29 & 99.41 & -- & 86.37 \\
			\hline
			VGG-16 &{\bf98.66}&	{\bf99.86} &  90.47	 & 89.71 \\
			ICNN &95.39& 96.51  & 89.11  &  {\bf91.60}\\
			compositional CNN & 97.12 & 99.27 & {\bf90.70} & 87.51\\
			\hline
			ResNet-18 & {\bf97.77} & {\bf99.81}  & 89.60  & --\\
			ICNN    & 93.30 & 97.12 & -- &  --  \\
			compositional CNN & 96.90 & 98.49 & {\bf89.76} &  --\\
			\hline
			ResNet-50    &  {\bf97.88}  &  {\bf99.88} & {\bf90.21} & --\\
			compositional CNN & 97.30 & 99.27 & 89.63 & --\\
			\hline
			DenseNet-121 & {\bf98.29}  & {\bf99.92} & -- & 91.28   \\
			ICNN    & 96.55 & 99.22 & -- & -- \\
			compositional CNN & 97.52  &  98.83 & -- & {\bf91.75}\\
			\hline
			DenseNet-161    & {\bf98.70}  & {\bf99.96} & -- & {\bf93.48}\\
			compositional CNN & 98.14 & 99.61  & -- & 92.66\\
			\bottomrule
	\end{tabular}}
	\caption{Comparisons of classification accuracy between ICNNs and compositional CNNs revised from different classic CNNs.}
	\label{table-acc}
\end{table}

\subsubsection{Curves of Inconsistency and Diversity} Note that, the two metrics of inconsistency and diversity were closely related. Generally speaking, the greater the diversity was, the lower the consistency was. Therefore, in order to fairly compare different CNNs' inconsistency of visual patterns under different diversity, we reported \emph{inconsistency-diversity} curves in this paper, as shown in Fig.~\ref{fig:metrics}. To this end, we sampled different values of {\small$\tau$} to obtain different pairs of \emph{inconsistency-diversity}, thereby obtaining \emph{inconsistency-diversity} curves. Given $n$ sampled values of $\tau$, $[\tau_1, \tau_2, \cdots, \tau_n]$, we could calculate $n$ pairs of \emph{inconsistency-diversity} values, $(p_1, q_1)$, $(p_2, q_2)$, $\cdots$, $(p_n, q_n)$. The sampling of $\tau$ was under the constraint that $q_1, q_2, \cdots, q_n$ were evenly distributed between $(0,1]$.

\begin{figure*}[tbp]
	\centering
	\includegraphics[width=\linewidth]{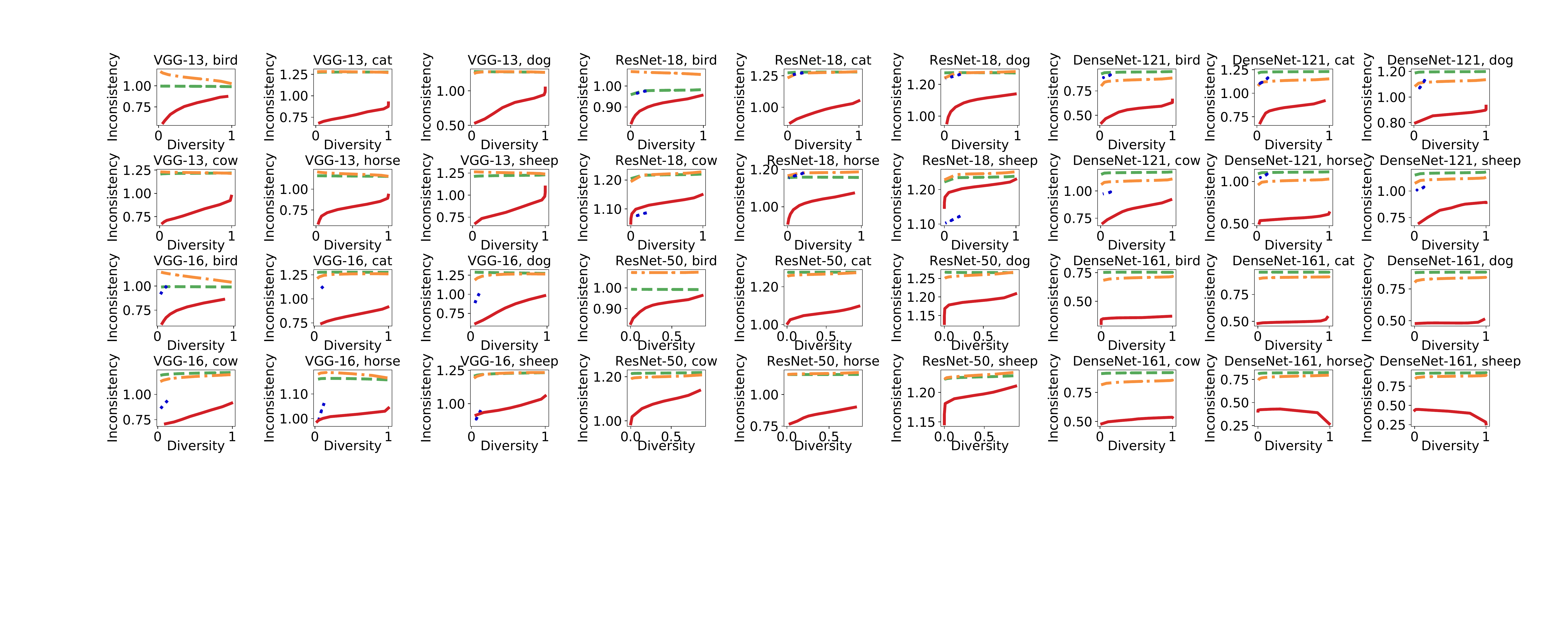}
	\includegraphics[width=0.6\linewidth]{fig2_legend}
	\caption{The \emph{inconsistency-diversity} curves of CNNs based on different categories of the PASCAL-Part dataset.}
	\label{fig:fig7}
\end{figure*}

\subsection{Experimental Results and Analysis}

\paragraph{\emph{Inconsistency-diversity} curves and classification accuracy.} Fig.~\ref{fig:metrics} shows the \emph{inconsistency-diversity} curves of different CNNs. Each inconsistency value was the average inconsistency over all filters. Under the same diversity of visual patterns, compositional CNNs exhibited higher consistency of visual patterns than traditional CNNs and ICNNs. Besides, compositional CNNs always showed higher diversity than ICNNs. As shown in Fig.~\ref{fig:fig2-fjq2} and Fig.~\ref{fig:mean_fmap}, interpretable filters of compositional CNNs could explain almost the entire region of the image, while filters of ICNNs could only represent small parts in ball-like areas. Note that sometimes we could not obtain large values of diversity for an ICNN, because RFs of all filters in the ICNN were small, as shown in Fig.~\ref{fig:metrics}. Traditional CNNs showed low consistency of visual patterns, which were close to that of randomly synthesized feature maps. This indicated that in terms of filter interpretability, features of filters in traditional CNNs did not show significantly better consistency than synthesized random features. As Table~\ref{table-acc} and Table~\ref{table:helen2} shows, compositional CNNs exhibited comparable classification performance with traditional CNNs. Besides, compositional CNNs achieved higher accuracy than ICNNs in most comparisons.

\begin{figure}[tbp]
	\centering
	\includegraphics[width=0.7\linewidth]{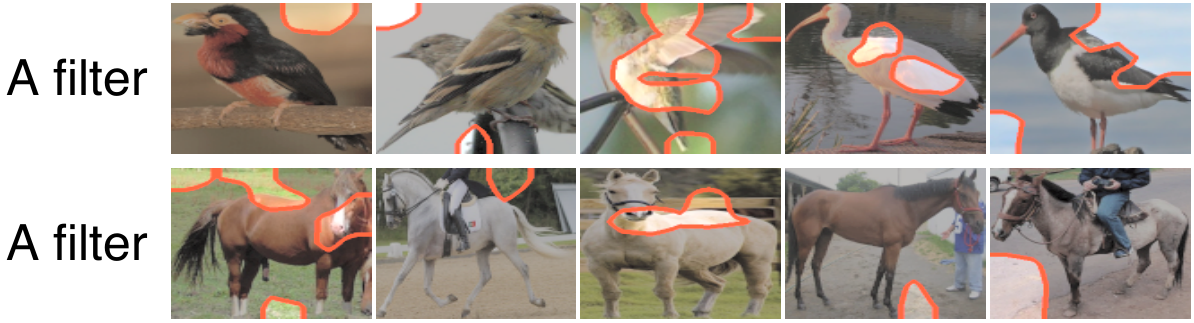}
	\caption{Very few cases when filters in compositional CNNs did not represent meaningful patterns.}
	\label{fig:failure}
\end{figure}

\begin{table}[tbp]
	\centering
	\resizebox{\linewidth}{!}{
		\begin{tabular}{c|cccccc}
			\toprule
			\!\!\!\!\!&\!\!\!\!	VGG-13 \!\!\!\!&\!\!\!\! VGG-16 \!\!\!\!&\!\!\!\! Res-18 \!\!\!\!&\!\!\!\! Res-50 \!\!\!\!&\!\!\!\! Dense-121 \!\!\!\!&\!\!\!\! Dense-161 \!\!\!\!\!  \\
			\midrule
			\!\!\!\!\! classic CNN & 1.0 & 1.0 &1.0 &1.0 & 1.0 & 1.0\\
			\!\!\!\!\! ICNN & -- & 99.70 & 1.0 & -- & 1.0 & -- \\
			\!\!\!\!\! compositional CNN & 1.0 & 1.0 & 99.85 & 1.0 & 99.85 & 1.0\\
			\bottomrule
	\end{tabular}}
	\caption{Classification accuracy of CNNs based on the Helen Facial Feature dataset. Res indicates ResNet; Dense indicates DenseNet.}
	\label{table:helen2}
\end{table}

\paragraph{Visualization of filters.} We followed \cite{zhang2018interpretable} to visualize RFs corresponding to a filter's feature maps. Fig.~\ref{fig:fig2-fjq2} shows RFs of features of compositional CNNs and ICNNs learned for the binary classification of a single category. In compositional CNNs, given different images, each filter consistently represented the same object part or the same image region. Different filters represented different object parts or image regions. In ICNNs, filters only represented small parts in ball-like areas. In addition, filters in the compositional CNN usually represented more complex shapes than filters in the ICNN. We specifically found out failure cases of interpretable filters in compositional CNNs, as shown in Fig.~\ref{fig:failure}. We also compared RFs between the center of the group and each filter in the group in Fig.~\ref{fig:group}.

\paragraph{Comparison of interpretable filters in different convolutional layers.} 
As shown in Fig.~\ref{fig:fig4} (a), filters of a high convolutional layer usually represented object parts or image regions, while filters of a middle convolutional layer usually represented local textures or local shapes.

\paragraph{Comparison of interpretable filters learned with different values of {\small$K$}.} As shown in Fig.~\ref{fig:fig4} (b), filters in the compositional CNN with {\small$K=16$} represented more visual patterns than filters in the compositional CNN with {\small$K=8$}.

\paragraph{t-SNE visualization.} We visualized filters in a compositional CNN and a traditional CNN using t-SNE \cite{maaten2008visualizing}. These two CNNs were learned based on the VGG-16 using the bird category in the PASCAL-Part dataset. As Fig.~\ref{fig:fig4} (c) shows, feature maps of a compositional CNN seem more clustered than those of a traditional CNN.

\section{Conclusion}

In this paper, we have proposed a method to modify a traditional CNN into a compositional CNN, in order to make filters in a high convolutional layer encode meaningful visual patterns without any part or region annotations for supervision. Specifically, we design a loss to encourage each filter in the layer consistently represents the same object part or the same image region through different images, and encourage different filters in the layer to represent different object parts and image regions. Experiments have demonstrated the effectiveness of our method.

\section*{Acknowledgments} This work is partially supported by the National Nature Science Foundation of China (No. 61976160, 61906120, U19B2043), and Shanghai Municipal Science and Technology Major Project (2021SHZDZX0102).

%\bibliographystyle{named}
%\bibliography{reference}

\end{document}